# Desktop-Delta Bench: Do Computer-Use Models Understand Desktop GUI Transitions?*

Abhishek Pillai†
NVIDIA
Austin, TX, USA

Samir Kumar Nayak
NVIDIA
Bangalore, Karnataka, IND

Yuan Chen
NVIDIA
Austin, TX, USA

## ABSTRACT

Computer-use agents (CUAs) increasingly act through desktop GUIs to complete long-horizon tasks. Current benchmarks primarily measure end-task success or single-frame grounding. Neither isolates whether a model can reconstruct the causal, task-relevant transition produced by an action- crucial for rejecting stale observations, verifying progress, and recovering from failure. This is difficult because inference, remote input, application rendering, and screenshot capture are asynchronous: the next observation may be delayed, occluded, transient, or unrelated, then misread as progress and carried into subsequent planning.

We introduce Desktop-Delta Bench (DDB), an offline step-level benchmark containing 2,013 human-verified instances from novel, multi-app Linux trajectories across ~15 applications and 50 task domains. DDB trajectories targets 3 failure dimensions- state verification, source tracking, and context-aware control- through 2 complementary tasks: 463 3-frame temporal-ordering instances, including 105 with a cross-trajectory decoy, and 1,550 before-after pairs labeled from 5 actions + its payload.

We evaluate 8 closed and open-source model families across 32 ordering and 16 single-action settings (48 total), observing consistent gaps. Ordering remains unsaturated: the best non-decoy and decoy exact-match rates are 65.1% and 65.7%. Task context improves decoy identification by 6.9 percentage points but reduces non-decoy exact match by 2.2 points, while error analysis reveals systematic copying of the presented A → B → C order. Single-action results show that inferring the action family is harder than locating it: click F1 reaches 0.96 but only 0.76 for drag, while recognized drags are generally localized well. DDB complements, rather than replaces, end-to-end benchmarks by filling the missing diagnostic layer between GUI grounding and final task success, enabling targeted improvements to desktop CUA verification, reliability, and recovery.

*We commit to releasing all task samples, evaluators, configuration, associated artifacts and VLM traces upon publication at: https://github.com/abhipi/DDB
†Corresponding Author

## 1 Introduction

Computer-use agents (CUAs) translate natural-language instructions into GUI actions on desktops. A CUA is a system: it couples a vision-language model (VLM) with an orchestration layer for tool calls, context windows, and action execution. Because inference, network, and rendering latencies decouple the agent's observation from the true GUI, the underlying model routinely receives non-deterministic input: focus changes, pop-ups, competing windows, and stale UI [1] [33]. End-to-end benchmarks such as OSWorld [2] and OSWorld 2.0 [3] score whether the agent completes a task in a live environment. This outcome-level signal is essential but attributes a single score to the composite system: a task failure may originate in the model's visual reasoning, in its context management, or in a non-deterministic environment outcome. Conversely, a passing score does not certify that the model maintained a correct state understanding throughout; the agent may have reached the goal despite an intermediate misinterpretation that happened not to be decision critical.

**Figure 1:** Desktop-Delta Bench's architecture. **(A)** Human demonstrators execute workflows on Ubuntu 24.04 (GNOME); a guest-side recorder captures frames at 10 fps with synchronized keyboard and mouse events, producing state trajectories $s_1$, $s_2$, … linked by actions $a_1$, $a_2$, …. Gold labels derive from recorded events. **(B)** Temporal ordering: 3 screenshots from 1 trajectory under shuffled opaque labels; the model selects 1 of 6 permutations. In 105/463 instances, 1 screenshot is a decoy from a different recording. Evaluated with and without task instruction; baseline 16.7%. **(C)** Single-action reconstruction: 2 frames (1 s before and after an action); model predicts action family (1/5) and structured payload. No task instruction provided.

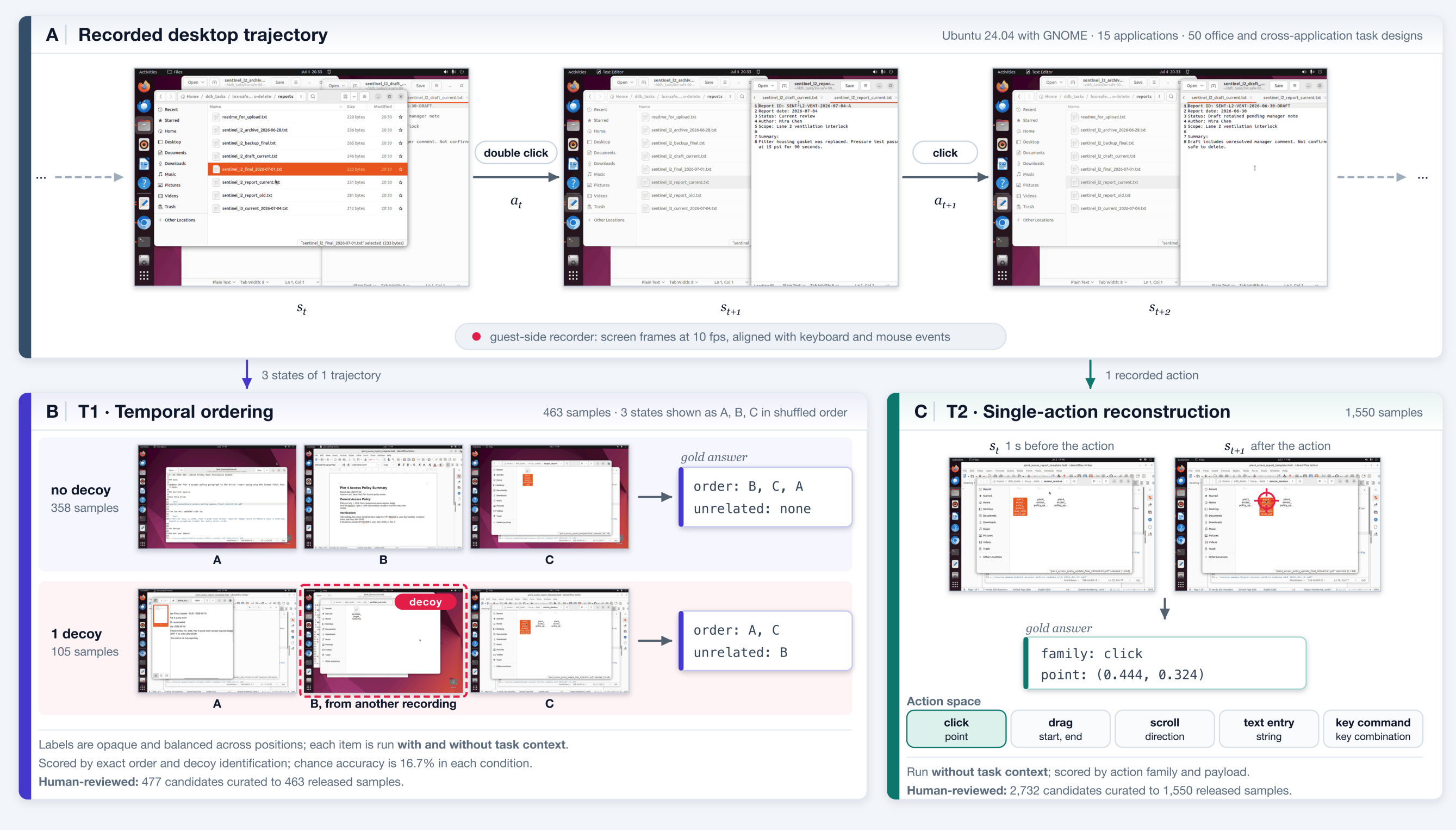

For CUAs, VLMs must interpret GUI state transitions, not just static screens. They must reconstruct chronology from a set of observations, infer what action produced an observed change, and reject visual evidence that does not belong to the active workflow. Failure to reject such "decoys" creates context poisoning: stale or misattributed desktop context enters the model's working context and biases subsequent reasoning. This risk is acute in desktop work, where multiple applications, files, and document versions can appear visually similar. It also carries safety implications: BLIND-ACT [4] demonstrates that agents continue executing under contradictory and infeasible conditions rather than halting, because the model fails to distinguish actionable state from noise

Several benchmarks evaluate model capabilities in isolation. ScreenSpot-Pro measures *grounding:* whether a model can localize an instructed UI element in a single screenshot [5]. UI-Vision scores next-action prediction given a goal, the current screen, and interaction history [6]. EvoGUI [7] evaluates temporal ordering, inverse action prediction, and successor selection over web trajectories from Mind2Web and WebLINX, establishing the value of transition-level model diagnosis at web scale. Native desktop and office workflows pose challenges that web trajectories do not: the model must verify hidden or persistent state changes (e.g., a save operation that alters no visible pixels), track the correct source across files and applications, and adapt its interpretation when active context shifts unexpectedly.

We introduce **Desktop-Delta Bench (DDB),** a benchmark that evaluates desktop state-transition understanding of VLMs on Linux through 2 complementary tasks. In **(1) temporal ordering**, the model reconstructs the chronology of 3 screenshots from 1 recording; in the decoy condition, it must additionally identify 1 screenshot drawn from a different recording, modeling the practical error in which a stale window, wrong file, or previous version enters context. In **(2) single-action reconstruction**, a model receives 2 frames immediately before and after one recorded action and predicts the action family (click, drag, scroll, text entry, or key command) together with its structured payload (click coordinates, drag endpoints, scroll direction, entered text, or key combination). This tests whether a model can recover actions from their visible effect, without any agent rollout.

DDB contains 463 temporal ordering (105 with decoy) and 1,550 single-action pairs from novel Ubuntu desktop recordings of daily office workflows across 50 task domains: document editing, spreadsheets, presentations, file management, email, and cross-application operations. **Collection targeted 3 failure dimensions:** state verification (hidden, persistent, and spatial changes), source and item tracking (files, versions, cross-item consistency), and context-aware control (rerouting, recovery, ambiguity, and stopping conditions). All released instances are human reviewed; we release a hard set that is entirely configurable in ordering permutations and domains for evaluation.

Our contributions are:

1. **Auditable Linux desktop benchmark:** 2,013 human-verified instances from office, cross-application workflows, released with golden labels, A/B/C maps, configurable evaluators, and 356 task artifacts.
2. **Failure diagnostic framework**: 2 offline prediction problems across 3 failure dimensions: (i) temporal ordering over 6 permutations of 3 screenshots with optional decoy rejection, scored on temporal accuracy and rejection rate; (ii) single-action reconstruction over 5 action families with structured payloads, scored on family precision, payload accuracy, and spatial error.
3. **Methodical analysis and validation:** 8 frontier closed and open-source models- with leading reported agent-level results on CUA benchmarks- evaluated across 48 configurations (ordering: 8 models × 2 inference settings × 2 prompt conditions, action: 8 models × 2 settings). This shows DDB is far from saturated.

**A U.S. provisional patent** covering the DDB model-evaluation technique for VLMs used in CUAs was filed on July 17, 2026, and is pending.

# 2 Background and Related Work

## 2.1 End to end computer use

OSWorld provides 369 tasks (101 multi-application) scored by execution-based endpoint evaluation in a live Ubuntu VM [Xie et al., 2024]. Its original evaluation reported 72.36% human accuracy and 12.24% for the best model. As of July 2026, frontier agents exceed 85% on OSWorld-Verified (https://osworld-v1.xlang.ai/), surpassing the human baseline. OSWorld 2.0 targets longer professional workflows: 108 tasks, median human completion time of ~1.6 hours, ~318 tool calls per agent [Yuan et al., 2026]. The best configuration achieves 20.6% binary completion and 54.8% partial score. Its documented failure modes- lost constraints, missed mid-task information, skipped verification, hidden-state recovery- directly motivate DDB's 3 failure dimensions. Windows Agent Arena [8] and WindowsWorld [9] report similar human-agent gaps, particularly in multi-application execution.

OSWorld-Human [10] measures agent efficiency relative to human step counts and completion times on the same tasks [Abhyankar et al., 2025]. AgentHijack [1] evaluates agent robustness to 9 environment corruptions (pop-ups, resolution changes, competing apps) on OSWorld tasks, motivating DDB's decoy and context-poisoning conditions. Both extend the OSWorld evaluation surface but retain end-to-end task completion as the scored unit. DDB uses the same style of workflows as source data but scores individual transitions.

## 2.2 GUI grounding

ScreenSpot-Pro [5] provides 1,581 instruction-element pairs from 23 professional desktop applications across 5 categories . OmniACT [11] annotates 9,802 UI elements across desktop and web UI-Vision [6] scores next-action prediction from 8,227 queries across 83 desktop applications [Nayak et al., 2025]. OSWorld-G evaluates fine-grained screenshot-conditioned manipulation [12]. All operate on single screenshots [20]. A predicted point can be spatially correct while the predicted event is semantically wrong- confusing right-click with left-click, selection with editing, or focused window. DDB scores action family and spatial location separately.

## 2.3 GUI transition and step-level learning

GUI-Shift [13] trains inverse dynamics on 2K AndroidControl [22][] GUI transition pairs, improving mobile. Computer-Using World Model predicts next state from a current desktop state and candidate action [14]. Both demonstrate that GUI dynamics are a useful learning signal. AgentProcessBench provides correct/neutral/error labels over tool-agent steps, showing that final success hides intermediate failures [15]. WebStep tracks deterministic semantic web states and shows that agents with similar endpoint success differ sharply in execution patterns [16][21][23]. BLIND-ACT finds that CUAs continue executing in 60-90% of ambiguous, unsafe, or infeasible conditions across 90 OSWorld-based tasks [4]. DDB includes trajectories around safety and should-stop decisions but scores their temporal structure and action mechanics, not refusal or policy behavior.

The closest comparison is EvoGUI [7], published recently (after our filing for a patent). It evaluates 3 transition-level tasks over 3,000 instances from Mind2Web [17] and WebLINX [18] web trajectories spanning 120 domains: temporal ordering (K = 3, 4, 5), inverse action/value prediction over click/type/select, and contrastive 1-step successor discrimination. DB differs on 6 axes:

- **Data source:** EvoGUI re-mines existing Mind2Web / WebLINX browser trajectories. DDB collects controlled Linux desktop recordings across 50 task domains.
- **Environment:** EvoGUI covers websites in a browser. DDB covers native applications, file choosers, office documents, terminals, and cross-application workflows.

| Benchmark | Domain | # Instances | Env. | Evaluates | Transition-level? |
|---|---|---|---|---|---|
| ScreenSpot-Pro [Li et al., 2025] | Desktop; 23 apps, 5 categories | 1,581 | Offline | Model: grounding | ✗ |
| OSWorld [Xie et al., 2024] | Desktop (Ubuntu); 9 apps | 369 | Live VM | Agent: task completion | *implicit* |
| OSWorld 2.0 [Yuan et al., 2026] | Desktop + 31 websites; 7 domains | 108 | Live VM | Agent: long-horizon completion | *implicit* |
| BLIND-ACT [Shayegani et al., 2025] | Desktop (Ubuntu); 9 apps | 90 | Live VM | Agent: safety | ✗ |
| UI-Vision [Nayak et al., 2025] | Desktop; 83 apps | 8,227 | Offline | Model: next action | ✗ |
| EvoGUI [Yang et al., 2026] | Web; 120 domains | 3,000 | Offline | Model: temporal order, inverse action, successor | ✓ web |
| **DDB (ours)** | **Desktop (Linux); 50 task domains, ~15 apps** | **2,013** | **Offline** | **Model: temporal order + decoy, 5-family action + payload** | ✓ desktop |

**Table 1:** Comparison of DDB with CUA-related benchmarks. **Domain:** platform and application coverage. **Instances:** number of scored tasks. **Env:** whether evaluation requires a live VM or runs offline over artifacts. **Evaluates:** whether the benchmark scores the VLM or agent. **Transition-level:** whether the benchmark explicitly tests state-transition understanding as opposed to single-screen grounding or end-to-end task completion where transition reasoning is only implicit.

- **Action families:** EvoGUI predicts click, type, or select with no coordinate reconstruction. DDB reconstructs 5 families with continuous click/drag coordinates, scroll direction, entered text, and key combinations.
- **Decoy design:** EvoGUI uses a separate binary successor task with an explicit distractor prompt. DDB **embeds decoys** in 105 ordering instances; the model is not told whether a decoy exists.
- **Failure dimensions:** EvoGUI uses trajectory-derived probes without targeted failure categories. DDB collection targets **3 dimensions:** state verification, source/item tracking, and context-aware control.
- **Curation:** EvoGUI automatically derives labels from trajectory logs. DDB uses human review to maintain quality (2,013/3,209 (62.7%) candidates retained).

A key difference is decoy handling. In real desktop workflows, they enter context without indication- DDB's design mirrors this by **not cueing the model to the presence of 50% irrelevant evidence.** EvoGUI establishes transition-level diagnosis for web. DDB extends this to Linux desktop workflows.

# 3 Method and Task Formulation

## 3.1 Desktop state and GUI-action mapping

Desktop state is not purely GUI. We represent latent state as:

$$s_t = (v_t,\ h_t,\ p_t,\ r_t,\ c_t)$$

where $v_t$ is visible GUI state, h_t is hidden application state (focus, selection, clipboard, dirty buffers), $p_t$ is persistent artifact state (files, saved documents, attachments), r_t is runtime/process state (alive, hung on OS), an $c_t$ is task context including corrections and safety constraints. A screenshot is a rendering:

$$o_t = R(s_t)$$

so distinct latent states can render almost identically (e.g., a file saved vs. unsaved may differ by a single title-bar character), and visible changes can be operationally irrelevant (e.g., a cursor blink). In most CUA patterns (including OS-World), the model does not infer on a video stream, but screenshots from agent-driven tool calls. Between 2 observations, several user, agent, or system events may have occurred, causing the desktop to change in ways unrelated directly to the last action. The observations $o_t$ and $o_{t+k}$ available to the model are therefore, in practice, separated by $k \geq 1$ unobserved intermediate states. This models real CUA conditions and **tests robustness to incomplete observation histories**- a failure mode that consecutive-frame web trajectory datasets do not expose. We employ this observation when framing the temporal ordering task (1).

## 3.2 Single Action Reconstruction (2)

Our recorder on the machine being executed on (the "target") yields an action-mapped trajectory through demonstrations:

$$\tau_{\text{map}} = \left\{(o_t,\ a_t,\ \tilde{o}_{t+1})\right\}_{t=0}^{T-1}$$

where $o_t$ is the event-aligned pre-action frame and $\tilde{o}_{t+1}$ is the reviewed frame at which the action's effect is evaluated. A screenshot captured after a single action may not clearly reflect that action's effect: application rendering delays, environment state changes (e.g., a notification appearing mid-action), I/O failures (e.g., an incomplete save), and concurrent background processes can all cause $\tilde{o}_{t+1}$ to diverge from the expected visual outcome. Pairs where $\tilde{o}_{t+1}$ did not unambiguously reflect $a_t$ **were excluded during human review.** Moreover, **multiple distinct actions can produce the same rendered transition.** Human review retains only ($o_t$,$\tilde{o}_{t+1}$) pairs for which 1 canonical action and its scored parameters are defensible from the 2 frames supplied. Thus, for **single-action reconstruction (2)**, each sample is a reviewed tuple:

$$d_i = (I_i^-,\ I_i^+,\ a_i,\ \theta_i)$$

where $I_i^-$ and $I_i^+$ are the pre-action and at-action frames, $a_i$ is the canonical action family, and $\theta_i$ is the family-specific payload. The model predicts:

$$f_\Delta(I_i^-,\ I_i^+) \rightarrow (\hat{a}_i,\ \hat{\theta}_i)$$

No task instruction is provided. Prediction is just from 2 frames.

The action space, payload and target are defined for DDB as:

| Family | Payload | Gold target |
|---|---|---|
| `click` | Normalized $(x, y)$ | Recorded click point |
| `drag` | Normalized start $(x_1, y_1)$ and end $(x_2, y_2)$ | Recorded drag endpoints |
| `scroll` | Cardinal direction | ∈ {up, down, left, right} |
| `text_entry` | Exact text string | Unicode NFC-normalized, whitespace-preserved |
| `key_command` | Ordered sequence of key combinations | Modifier order: CTRL, ALT, SHIFT, META + base key |

**Table 2:** Payload specification for the 5 action families in single-action reconstruction (2).

## 3.3 Temporal Ordering (1)

As noted in section 3.1, agent actions in quick succession or concurrent desktop environment events (notifications, focus changes, background process completion), may cause consecutive agent-driven observations to diverge from the expected action-effect sequence. The 3 screenshots in each ordering triple are therefore not clean, consecutive recording frames- they are sampled from action-mapped positions across the trajectory, separated by seconds or minutes of workflow with many intermediate actions unobserved. The model must reconstruct chronology from high-level state progression (e.g., a document going from empty to edited to saved-and-closed), not from incremental pixel differences.

Each sample contains 3 screenshots under opaque labels $O = \{A, B, C\}$. Let $R \subseteq O$ denote options from the same source trajectory and $U = O \setminus R$ denote unrelated options. The model predicts an unrelated set $\hat{U}$, an induced related set $\hat{R} = O \setminus \hat{U}$, and a chronological ordering over $\hat{R}$:

$$f_{\text{ord}}(I_A,\ I_B,\ I_C,\ [q]) \rightarrow (\pi(\hat{R}),\ \hat{U})$$

where task instruction $q$ is absent in the minimal condition and present in the task-conditioned condition. Moreover, 2 variants:

- Same-trajectory ordering: $|R| = 3, |U| = 0$
- Ordering with decoy rejection: $|R| = 2, |U| = 1$

Each variant has 6 valid outcomes ( $3! = 6$ for same trajectory; $\binom{3}{1} \times 2! = 6$ for decoy). **The model is not told which condition applies.** Thus, under $|U| \in \{0,1\}$, the mixed output space has 12 outcomes. We report baseline at 8.3% (1/12, condition-unaware). VLMs recognize chronological presentation (A, B, C) trivially- the task reduces to confirming the given order rather than reconstructing it from visual evidence. We shuffle all triples for even distribution.

We pick 3 frames in this task, as 2 frames yield a binary (50%) ordering decision with no room for a decoy, as it would appear in practice. 4+ frames grow the output space factorially ( $4! = 24$ , $5! = 120$ ), increase image-token cost, and make unique-order auditing impractical. CUA patterns from model providers [19], and benchmarks like OS-World also majorly adopt 3 screenshots. 3 is the minimum that requires transitive consistency across 3 pairwise relations, supports a meaningful 1-decoy design, and keeps baseline measurable. **Every triple must have exactly 1 defensible chronological order from the frames.** Ambiguous triples are discarded during human review.

# 4 Benchmark Construction

## 4.1 Environment and Task Distribution

All 50 task domains were recorded on an Ubuntu 24.04 LTS x86_64 VM running GNOME at 4 native resolutions: 1176×768, 1920×1080, 1024×768, and 1304×848. Coordinates are stored as both raw pixels and normalized [0, 1] values, so model predictions are resolution independent (EvoGUI finds that resolution does not change transition prediction failures). ~15 office-task applications are used. Browser workflows use local `file://` portals, requiring no network or live user accounts. Each recording targets a dominant diagnostic failure from the 3 dimensions introduced in Section 3. Detailed sample and application distribution are shown in figures 2, 3, and 4. Individual recordings details with its dimension map are in Table A3 in Appendix A.

**Figure 2** Distribution of the 463 ordering triples by failure dimension, task type, and same-trajectory vs. decoy condition.

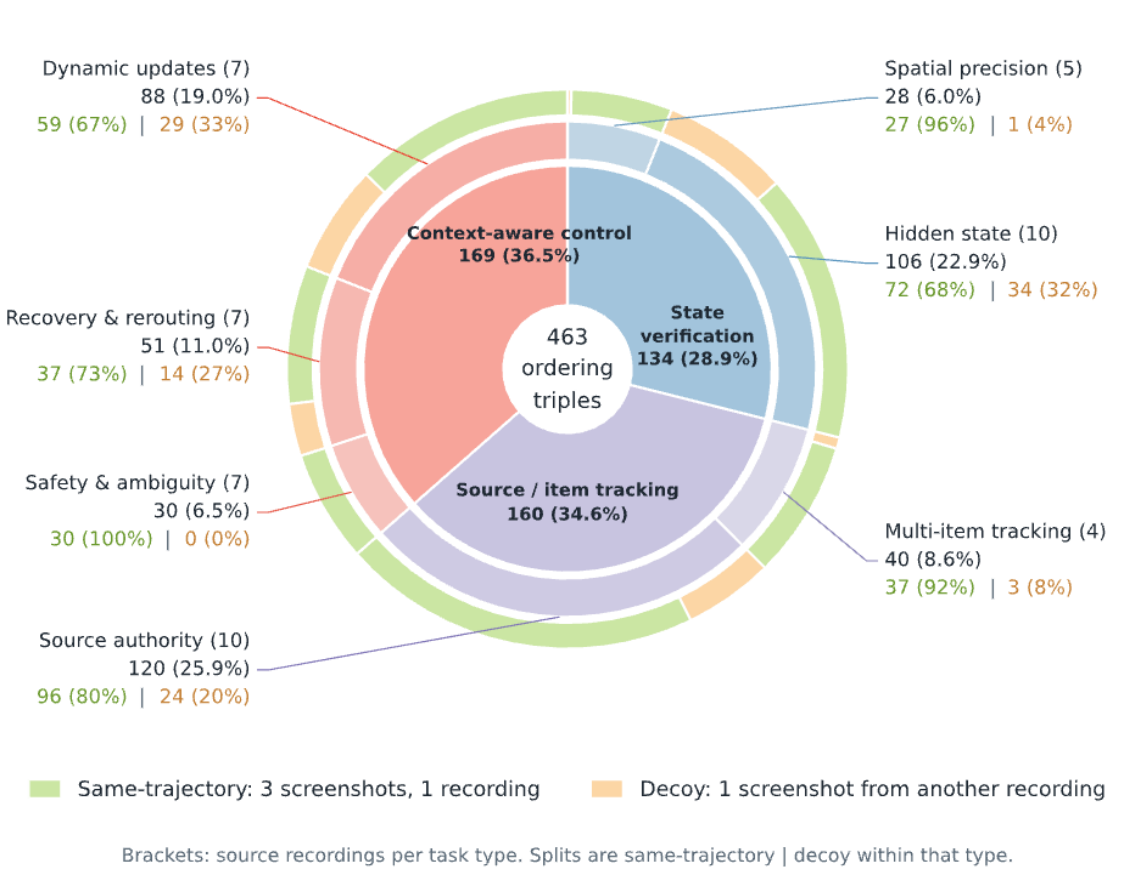

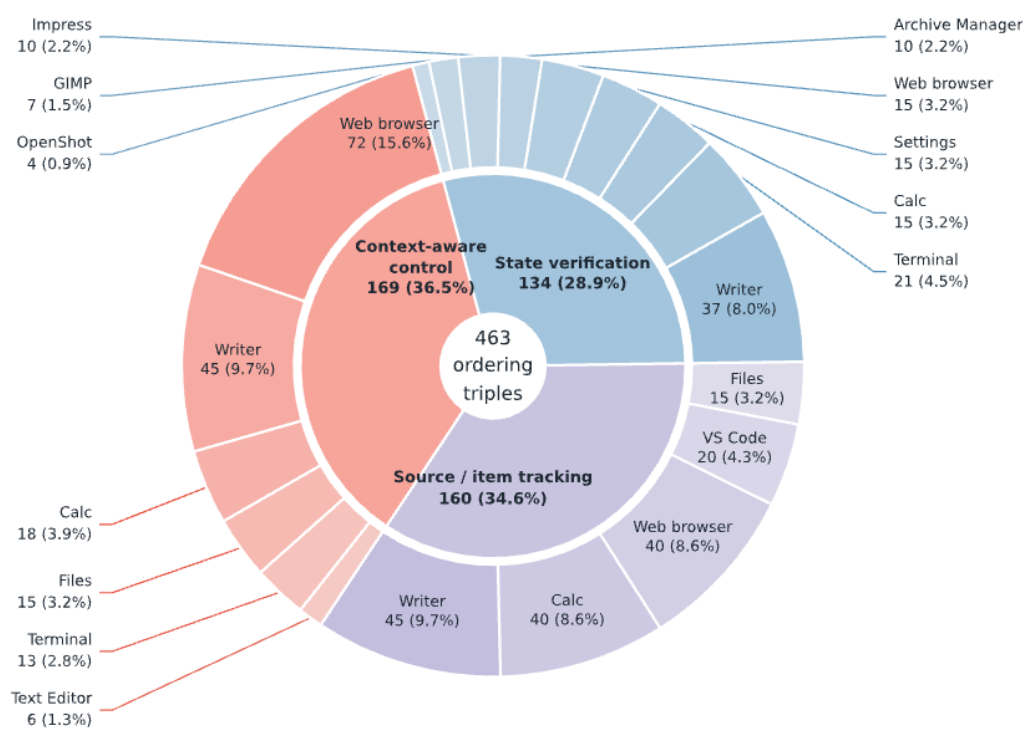

**Figure 3:** Application distribution of the 463 ordering triples across the 3 failure dimensions, placed by focused app for sample.

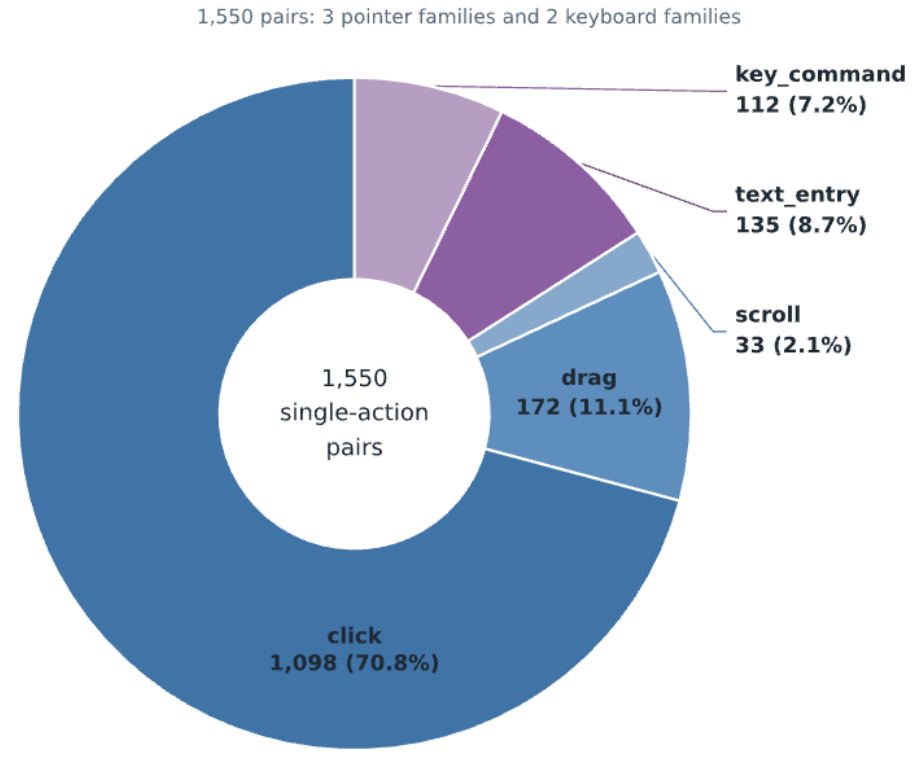

**Figure 4:** Action distribution of single-action reconstruction task.

The 1,550 single-action samples are dominated by clicks (70.8%), followed by drags (11.1%), text entry (8.7%), key commands (7.2%), and scrolls (2.1%). This reflects the natural distribution of office desktop workflows rather than a balanced design choice. Clicks outnumber other families by ~6×, so we report macro-F1 over the 5 action families as the primary metric.

## 4.2 Recording Pipeline

A Python pynput-based recorder on the target VM captures mouse and keyboard input while a parallel process records the screen at 10fps. The pipeline aligns these into the action-mapped trajectory to $\tau_{map}$ defined in Section 3, producing per-event before / after frame pairs and coordinate metadata. Each event defines a 1-second execution window. before.png is the last frame before the window opens; after.png is the first frame after it closes.

- **Drags**: mouse-down to up plus a 1s settling buffer.
- **Text entry**: keystrokes within 1s are chunked as 1 action
- **Key combinations** (e.g., Ctrl+S): modifier-down through base-key-up forms 1 action

The recorder logs fine-grained event types (mouse.singleclick, mouse.doubleclick, mouse.tripleclick, mouse.drag, mouse.scroll, key.type). During staging, these are normalized into the 5 canonical families of Table 2: single-, double-, and triple-clicks collapse into click; key.type events split into text_entry (printable character sequences) or key_command (special keys, shortcuts, modifier combinations) based on content; drag and scroll map directly. Key tokens are canonicalized via the alias table in Table 2 (e.g., left → ARROWLEFT, ctrl → CTRL).

A validator checks raw-to-normalized coordinate consistency within tolerance $5.1 \times 10^{-7}$. This produced 3,209 candidates: 477 ordering triples and 2,732 single-action pairs.

## 4.3 Human Review for Quality

Every candidate mined by one collector, receives 1 of 3 statuses from the second: approved, corrected, and indeterminate (excluded as ambiguous or can be mapped to ≥ action). This process reduced the ordering candidates from 477→463 and single-action from 2732→1550. Total: **3209→2013 (62.7%).**

The **final benchmark** contains folders:

- Ordering: 463 triples (A.png, B.png, C.png, label.json)
- Single-action: 1550 pairs (before/after.png, label.json)
- Artifacts: 356 files across 50 task directories

# 5 Evaluation Metrics

## 5.1 Label Balancing and Prompt Conditions

VLMs can echo the input label order (predict A→B→C) rather than reason over visual progression. To control this, the evaluator reassigns A/B/C label permutations evenly (default seed 42, configurable: 59-60 assignments each for same-trajectory triples (n=358); decoy slot distributed 34-36 times across the 3 positions (n=105). The input-order preservation rate- fraction of incorrect predictions that echo A→B→C is tracked per model.

Temporal ordering runs under 2 conditions: minimal (images only) and task-conditioned (task instruction prepended). Single-action reconstruction is task-independent only: the model sees 2 frames with no instruction. Exact context layout is in Appendix B.

## 5.2 Temporal Ordering Metrics

1. **Exact-match accuracy (primary):** A prediction is correct iff it matches the gold chronological order *and* the gold unrelated set. Duplicates, missing, invalid responses score 0.
2. **Prompt Condition and Decoy splits:** Separate accuracy for same-trajectory and decoy, minimal vs. task-conditioned.
3. **Pairwise accuracy:** Partial-credit metric: for each gold ordered pair among related states, we check whether the prediction preserves that precedence. Same-trajectory triples contribute 3 pairs; decoy triples contribute 1 pair.
4. **Baselines:** Condition-aware uniform random: 1/6 (16.7%). Condition-unaware: 1/12 (8.3%, primary) (Section 3).
5. **Trivial ordering:** Bias through input-order preservation rate (A→B→C echoed) and full reversal rate (C→B→A).

## 5.3 Single-Action Reconstruction Metrics

Scoring targets: family classification, non-spatial payload, and spatial reconstruction.

1. **Family classification:** Macro F1 (primary) $= \frac{1}{5}\sum_{c=1}^{5} F_{1,c}$: unweighted mean of per-class F1 over the 5 families. Click comprises 70.8% of samples; macro F1 prevents click from dominating**.** Per-family F1 is reported alongside to compare,
2. **Non-spatial payload (gold-family samples only):** Scroll: exact cardinal direction match. Text entry: exact normalized text match. Key command: exact normalized, ordered key-sequence match, alias-normalized per Table 2 (27 aliases, e.g., CMD/COMMAND → META). End-to-end accuracy per family: correct family ∧ correct payload.

*5.3.1 Spatial Grounding Metrics:* Point error is diagonal-normalized. Given predicted $(p_x, p_y)$ and gold $(g_x, g_y)$ in normalized [0, 1] coordinates at resolution $W \times H$ px:

$$e(p,g) = \frac{\sqrt{((p_x - g_x)\,W)^2 + ((p_y - g_y)\,H)^2}}{\sqrt{W^2 + H^2}}$$

For drag, joint error $e_{\text{drag}} = \max(e_{\text{start}}, e_{\text{end}})$. End-to-end accuracy at threshold $t$:

$$\mathrm{Acc}(t) = \frac{|\{i : \hat{a}_i = a_i \wedge e_i \le t\}|}{N_{\text{family}}}$$

where $N_{\text{family}}$ is the full gold-family population; misclassified samples count as failures at every threshold. We sweep $t$ over [0%, 20%] of screen diagonal (201 thresholds, 0.1 pp steps) and report success at 1%, 2%, 3%, 5%, 10%, 20% margins. The 20% upper bound covers a wide margin: many desktop actions (e.g., window focus) lack precise bounding boxes, and **DDB does not score pure grounding.** The resulting curves support the model-comparison analysis in Section 6. This can be summarized with normalized area-under-curve (AUC):

$$\mathrm{AUC} = \frac{1}{t_{\max}} \int_0^{t_{\max}} \mathrm{Acc}(t)\, dt, \quad t_{\max} = 0.20$$

Since AUC depends on correct family classification, we also track Action + Payload accuracy (correct family ∧ every scored field correct) to measure overall success coverage.

# 6 Experiments

## 6.1 VLMs and Inference Configuration

We evaluate 8 VLMs guided by the similarly top-performing reported entries on OSWorld (V1) and OSWorld 2.0 (reported score on V1 noted in parentheses in the following statement). 4 closed-source models were accessed via provider APIs: GPT 5.6-Sol [24] (5.5 scored 78.7%), Claude Opus 4.8 [25] (82.3%), Claude Sonnet 5 [26] (81.2%), and Gemini 3.6 Flash [27] (83%). 4 open-source models were benchmarked alongside: Holo-3.1-35B-A3B [28] (80.4%, Qwen 3.5 based), MiniMax M3 [29] (75.2%), Kimi K2.6 [30] (73.1%), and GLM-4.6V [31]. Locally hosted models are run in BF16 precision with vLLM inference [32]. All in the same configuration receive the same prompt and JSON parsing evaluator, decoded at temperature 1 (recommended). All are tested with reasoning enabled (provider managed/native) except Kimi-K2.6, in 2 settings each: medium/high for closed-source, 2048/4096 reasoning token budgets for open-source models (2K/4K). Each sample is attempted upto 3 retries, with no-cache reuse and a 600s timeout, and the best performance is retained for final scoring. This, in total, produces 32 ordering (8 models × 2 inference settings × 2 prompt conditions) and 16 single-action configurations (8 models × 2 inference) that we report (48 total). Closed models were accessed between July 15-July 25, 2026.

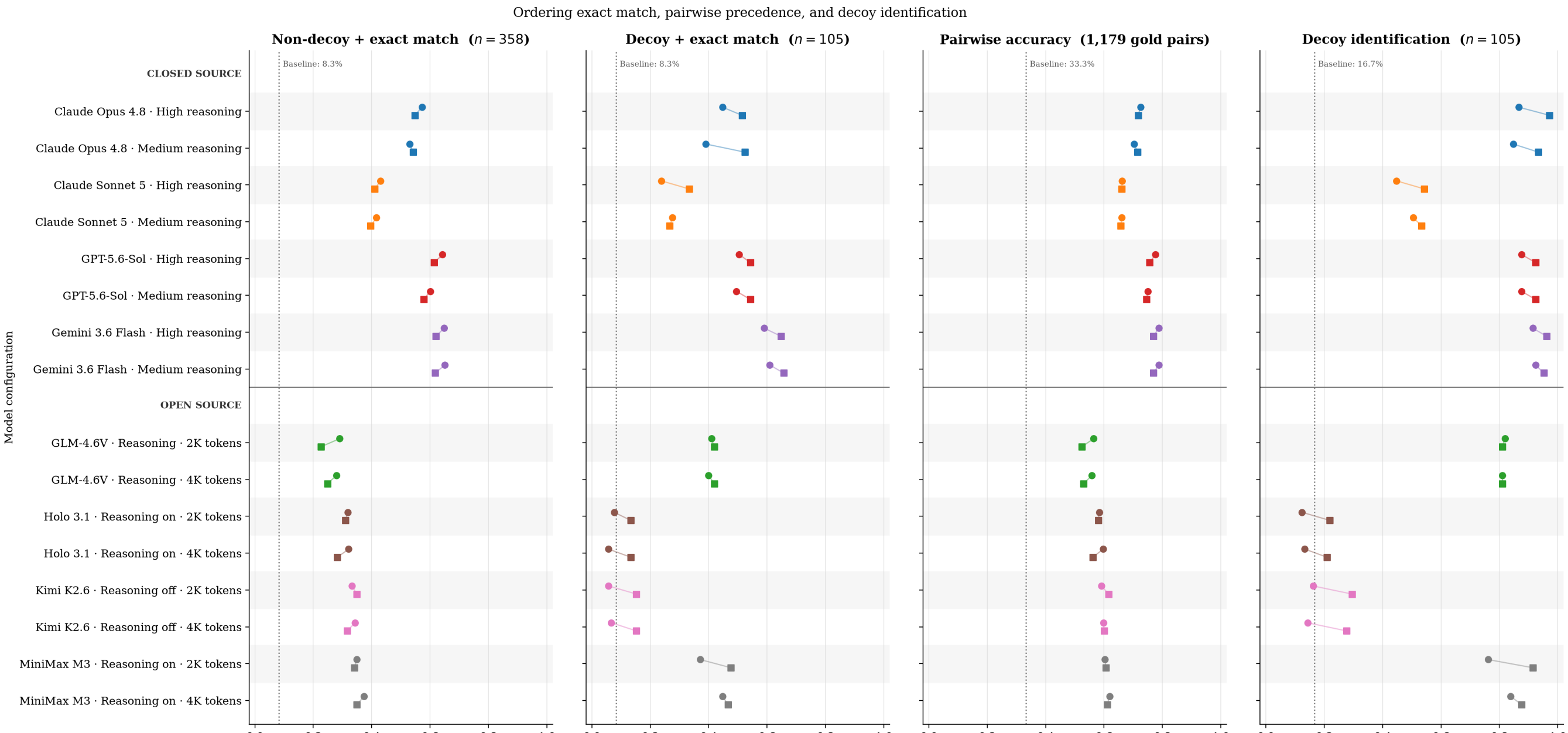

**Figure 5:** Temporal ordering results across 32 settings (16 model configurations × 2 prompt conditions). Panels decompose ordering into non-decoy exact match (n=358), decoy exact match (n=105), pairwise accuracy (1,179 gold pairs), and decoy identification (n=105). Circles: minimal prompt; squares: task-conditioned. Dashed lines mark condition-unaware random baselines (Section 3.3). Task context raises decoy identification (+6.9 points mean) while lowering non-decoy exact match (−2.2 points), helping trajectory membership not chronology. Closed-source configurations average 53.5% exact match versus 30.6% open-source (+22.9 points), but the pairwise gap narrows to 14.2 points. Claude Opus 4.8 reaches 97.1% decoy identification at only 51.4% decoy exact match; Holo 3.1 and Kimi K2.6 hold 56-62% pairwise while decoy identification falls to 12-30%. Detailed metrics in **Table A1 (Appendix A).**

## 6.2 Temporal Ordering Results

Figure 5 summarizes ordering performance by percentage points (pp) across all 32 settings. Per-configuration values for every metric appear in Table A1 (Appendix A). Metric definitions and baselines are in Section 5.

*6.2.1 Accuracies and Task Exposure:* Non-decoy exact match spans 22.6-65.1%, decoy exact match 5.7-65.7%, pairwise precedence 52.4-78.8%, and decoy identification 12.4-97.1%. Gemini 3.6 Flash attains the highest non-decoy exact match (65.1%, medium, minimal), decoy exact match (65.7%, medium, task), and pairwise accuracy (78.8%, both settings, minimal); Claude Opus 4.8 the highest decoy identification (97.1%, high, task). The lowest are GLM-4.6V (22.6% non-decoy exact match; 52.4% pairwise) and Holo 3.1 (5.7% decoy exact match; 12.4% decoy identification). Averaged across all 16, task conditioning raises decoy identification (6.9 pp) and decoy exact match (5.9) and lowers non-decoy exact match (-2.2). Thus, **task context improves trajectory membership, not temporal ordering. Reasoning affects performance minimally (all $|\Delta| \leq 4.5$ pp)**

*6.2.2 Closed vs Open Source:* Closed-source configs average 53.5% overall exact match against 30.6% open-source under the minimal prompt (22.9 pp; 23.2 under task). MiniMax M3 matches Claude Sonnet 5 on overall exact match (39.1% versus 39.3%) and exceeds it on both decoy metrics, but the group gap holds on all metrics (14.2-30.8 pp; Table A1).

*6.2.3 Trivial Ordering Failures:* Presentation-order echo is concentrated in 2 models: across 8 GLM-4.6V and MiniMax M3 configurations, **A → B → C accounts for 37.7-45.6% of failures,** compared with 1.6–5.5% for C → B → A (Figure 6).

| | MINIMAL | | | TASK | | |
|---|---|---|---|---|---|---|
| | Failures | A→B→C | C→B→A | Failures | A→B→C | C→B→A |
| CLOSED SOURCE | | | | | | |
| Claude Opus 4.8 · High reasoning | 211 | 43 (20.4%) | 19 (9.0%) | 213 | 42 (19.7%) | 16 (7.5%) |
| Claude Opus 4.8 · Medium reasoning | 232 | 47 (20.3%) | 19 (8.2%) | 214 | 42 (19.6%) | 17 (7.9%) |
| Claude Sonnet 5 · High reasoning | 284 | 53 (18.7%) | 45 (15.8%) | 281 | 52 (18.5%) | 39 (13.9%) |
| Claude Sonnet 5 · Medium reasoning | 285 | 56 (19.6%) | 53 (18.6%) | 293 | 52 (17.7%) | 45 (15.4%) |
| GPT-5.6-Sol · High reasoning | 180 | 29 (16.1%) | 21 (11.7%) | 186 | 30 (16.1%) | 22 (11.8%) |
| GPT-5.6-Sol · Medium reasoning | 196 | 31 (15.8%) | 25 (12.8%) | 199 | 25 (12.6%) | 29 (14.6%) |
| Gemini 3.6 Flash · High reasoning | 169 | 29 (17.2%) | 22 (13.0%) | 173 | 28 (16.2%) | 24 (13.9%) |
| Gemini 3.6 Flash · Medium reasoning | 166 | 24 (14.5%) | 22 (13.3%) | 173 | 22 (12.7%) | 24 (13.9%) |
| OPEN SOURCE | | | | | | |
| GLM-4.6V · Reasoning · 2K tokens | 316 | 119 (37.7%) | 5 (1.6%) | 338 | 154 (45.6%) | 17 (5.0%) |
| GLM-4.6V · Reasoning · 4K tokens | 321 | 123 (38.3%) | 12 (3.7%) | 330 | 141 (42.7%) | 12 (3.6%) |
| Holo 3.1 · Reasoning on · 2K tokens | 341 | 86 (25.2%) | 27 (7.9%) | 338 | 109 (32.2%) | 20 (5.9%) |
| Holo 3.1 · Reasoning on · 4K tokens | 342 | 75 (21.9%) | 35 (10.2%) | 348 | 115 (33.0%) | 20 (5.7%) |
| Kimi K2.6 · Reasoning off · 2K tokens | 338 | 72 (21.3%) | 24 (7.1%) | 322 | 71 (22.0%) | 28 (8.7%) |
| Kimi K2.6 · Reasoning off · 4K tokens | 333 | 61 (18.3%) | 34 (10.2%) | 334 | 66 (19.8%) | 24 (7.2%) |
| MiniMax M3 · Reasoning on · 2K tokens | 299 | 125 (41.8%) | 16 (5.4%) | 291 | 129 (44.3%) | 15 (5.2%) |
| MiniMax M3 · Reasoning on · 4K tokens | 282 | 117 (41.5%) | 14 (5.0%) | 289 | 130 (45.0%) | 16 (5.5%) |

**Figure 6:** Ordering failure output bias. Permutation balanced by seed (Section 5). Cell shading encodes magnitude within column.

## 6.3 Single-Action Reconstruction Results

Table 2 reports the best reasoning setting per model; Figures 7 and 8 gives spatial tolerance curves (click, drag), and per-configuration values appear in Table A2 (a/b) (Appendix A).

**Table 2:** Single-action reconstruction using each model's best reasoning effort. Columns report Macro F1, per-family F1, and exact-payload accuracy (%) for scroll, text, and key actions. Bold and underlined values denote best and second best per column; header values show the majority-click baseline.

| Model | Macro F1 | Click | Drag | Scroll | Text | Key | $Scr_{ex}$ | $Txt_{ex}$ | $Key_{ex}$ |
|---|---|---|---|---|---|---|---|---|---|
| *Majority baseline* | *0.166* | *0.83* | *0.00* | *0.00* | *0.00* | *0.00* | — | — | — |
| *Closed-source* | | | | | | | | | |
| Gemini 3.6 Flash | **0.800** | **0.96** | **0.75** | **0.70** | 0.93 | **0.66** | 84.8 | 81.5 | **46.4** |
| Claude Opus 4.8 | 0.776 | 0.96 | 0.73 | 0.61 | 0.93 | 0.66 | 87.9 | **84.4** | 37.5 |
| GPT-5.6-Sol | 0.756 | 0.93 | 0.67 | 0.63 | **0.95** | 0.61 | **90.9** | 66.7 | 33.0 |
| Claude Sonnet 5 | 0.754 | 0.95 | 0.60 | 0.67 | 0.93 | 0.62 | 84.8 | 74.1 | 38.4 |
| *Open-source* | | | | | | | | | |
| MiniMax M3 | 0.633 | 0.91 | 0.40 | 0.62 | 0.89 | 0.35 | 87.9 | 49.6 | 15.2 |
| Kimi K2.6 | 0.562 | 0.91 | 0.38 | 0.39 | 0.84 | 0.29 | 81.8 | 52.6 | 8.0 |
| Holo3.1-35B-A3B | 0.537 | 0.90 | 0.45 | 0.43 | 0.77 | 0.14 | 48.5 | 38.5 | 2.7 |
| GLM-4.6V | 0.476 | 0.85 | 0.38 | 0.34 | 0.58 | 0.23 | 45.5 | 43.0 | 9.8 |

The 4 closed-source models average 0.768 macro F1 (0.748-0.800) against 0.550 (0.476-0.633) open-source, a 21.8pp gap. Moreover, Click F1 spans 0.85-0.96 and click AUC reaches 0.857, but no other family approaches this. Text, a direct GUI change, is next (0.95 best) but has high variance, falling to 0.77 for Holo3.1-35B-A3B and 0.58 for GLM-4.6V; scroll 0.70 on Gemini 3.6 Flash against 0.34 for GLM-4.6V. Drag localization also lags click by a wide margin. AUC spans 0.552-0.857 for clicks but only 0.106-0.535 for drags (Figures 7, 8). The gap is smallest for Gemini 3.6 Flash (0.846→0.529; 1.6×) and largest for Kimi (0.850→0.196; 4.3×) and MiniMax (0.633→0.119; 5.3×).

Recognizing the family is far easier than reconstructing the exact payload. Text entry reaches 84.4% for Claude Opus 4.8 but 38.5% for Holo3.1-35B-A3B and scroll 90.9% for GPT-5.6-Sol against 45.5% for GLM-4.6V, when it only has 4 cardinals. Key command is unsaturated: Gemini 3.6 Flash leading at 46.4%. Such step-level analysis and failures track gaps uncovered by end-to-end benchmarks, directing meaningful VLM improvement.

**Figure 7:** Click spatial tolerance. Click accuracy versus localization error margin (0-20% of the frame diagonal). Family misclassifications fail at all margins. **Higher, to the left is better.**

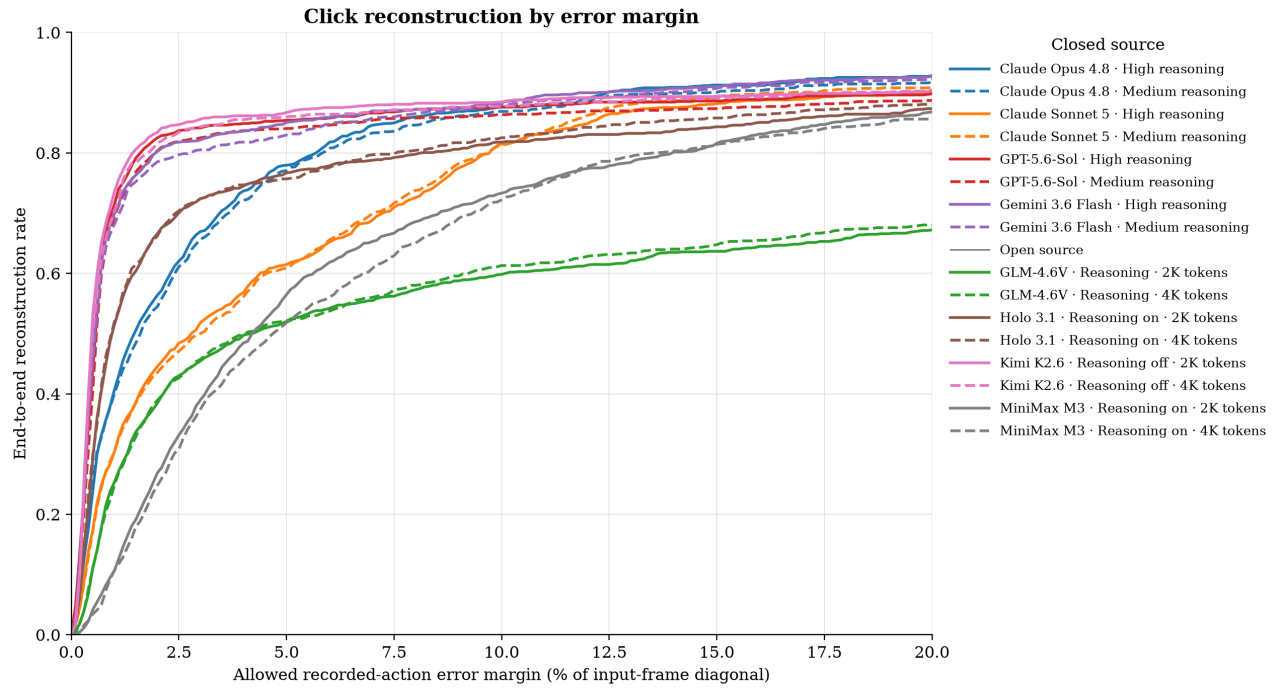

**Figure 8:** Drag spatial tolerance. Drag accuracy versus joint endpoint error margin (0-20% of the frame diagonal). Family misclassifications fail at all margins. **Higher, to the left is better.**

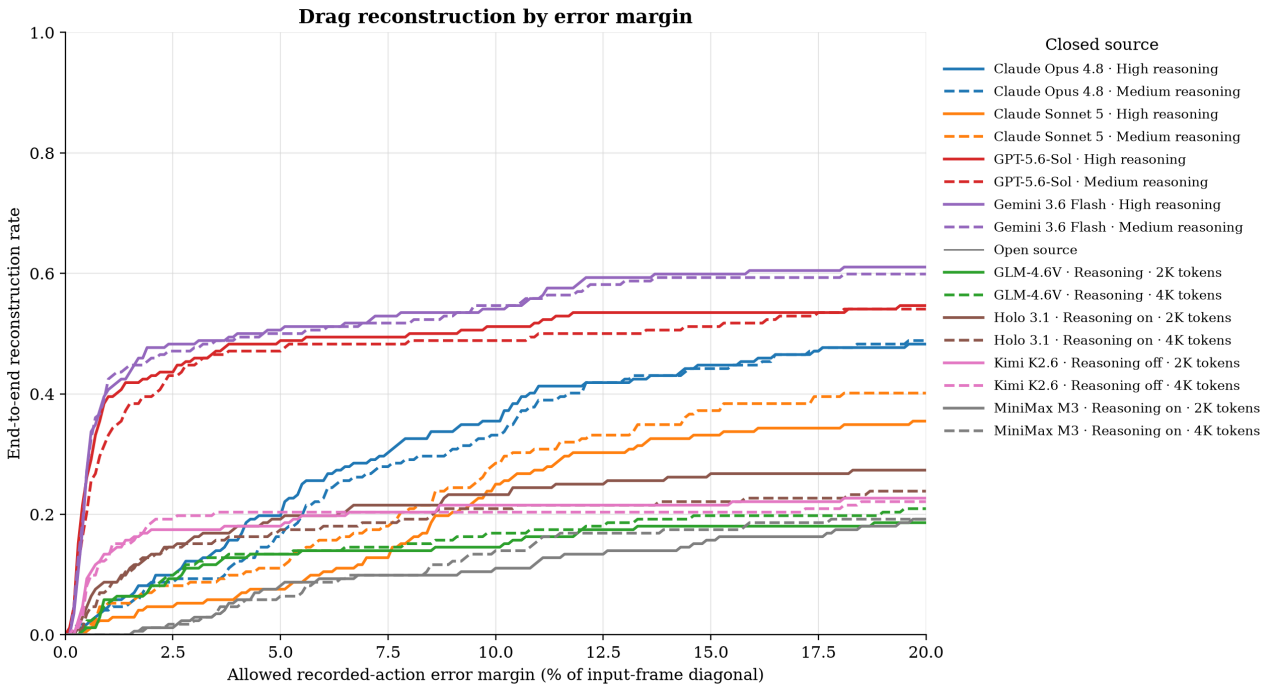

# 7 Conclusion

We evaluate 14,816 ordering predictions and 24,800 single-action predictions. **Ordering remains far from solved:** the best configuration reaches 65.1% non-decoy exact match, 65.7% decoy exact match, and 78.8% pairwise, even though decoy identification reaches 97.1%. Adding the task instruction helps identify trajectory membership but not chronology; reasoning has marginal impact. Among GLM-4.6V and MiniMax M3 failures, 37.7-45.6% preserve the presented A → B → C order, compared with only 1.6-5.5% that produce the reverse C → B → A order. This reveals a **presentation-order copying bias,** that will go unflagged without GUI-transition level analysis. **Single-action results show that inferring which action occurred can be harder than localizing it.** These gaps provide concrete improvement targets in action and payload recovery. DDB is not intended to replace end-to-end benchmarks; it is the missing step-level diagnostic layer between GUI grounding and final task success, separating family, payload, and spatial failures so improvements can be targeted toward more reliable CUAs.

# 8 Limitations

DDB's 2,013 offline instances are across ~15 office-oriented Linux applications; extending to Windows and macOS would test transfer across window management, shortcuts, and dialogs. Its 5 action families cover primitives, while application-specific tools and broader CUA action spaces motivate new labels and recordings. Finally, 1-action pairs and 3-state orderings isolate step-level understanding but do not measure long-horizon context accumulation: longer sequences and closed-loop episodes are important extensions.

# 9 Acknowledgments

We thank our colleagues at NVIDIA- Joe Li, Nat Duca, Pascal Berard, and Michael Fu- for discussions and internal reviews that helped refine benchmark design and evaluation metrics. We are grateful to the inference infrastructure team for supporting the VLM-evaluation runs. We also acknowledge the developers/maintainers of OSWorld, Ubuntu, GNOME, LibreOffice, VS Code, and other open-source software used.

# 10 Ethical Considerations

DDB uses novel, purpose-recorded workflows performed by human demonstrators in controlled, clean Ubuntu environments. Released screenshots, actions, and labels are human-reviewed; credentials and unintended sensitive content are excluded or redacted. Evaluation is offline and cannot execute predicted actions. DDB's purpose is for CUA research, not user monitoring.

# APPENDIX

The appendix provides the specifics underlying the main results. Appendix A reports all evaluated configurations, analyzes temporal ordering by trajectory failure dimension, maps exact recordings and samples to these dimensions, and documents the human review tools constructed for label review. Ordering uses 463 unique samples per prompt condition (358 same-trajectory and 105 decoy); single-action reconstruction uses 1,550 samples per configuration.

## A Results, Recording Map, and Human Review

- A.1: Temporal Ordering
  - Table A1: 32 configurations and 5 metrics.
  - Figure A1-a: Mean accuracy by trajectory failure.
  - Figure A1-b: Task-conditioning gains and losses.
- A.2 : Single-Action Reconstruction
  - Table A2-a: Macro and per-family F1.
  - Table A2-b: Click/drag AUC and exact scroll/text/key recovery.
- A.3: Task-level coverage and trajectory-failure map
  - Table A3: 50 task summaries, trajectory failures, dimensions, applications, and sample counts.
- A.4: Human review and correction interfaces
  - Figure A4-a: Ordering-review interface
  - Figure A4-b: Single Action-review interface

## B Task Context and Reproducibility Parameters

- B.1: Minimal and Task-conditioned ordering prompts
- B.2: Single-Action prompt and output schema
- B.3 Evaluation Seed and Reasoning Settings

## A.1 Temporal Ordering Results

**Table A1:** Results across 32 configurations (8 models × 2 reasoning settings × 2 prompt conditions). Percentage values over 463 samples. Min. = images + minimal prompt; Task = task instruction prepended. Each row has N = 463 (358 same-trajectory, 105 decoy). Overall exact is strict exact match across all samples. Pairwise is computed over gold precedence pairs, Decoy ID (identification) scores the unrelated label only. Within each prompt condition and metric, best is bold and second best is underlined: per column and per reasoning effort; ties share rank, using unrounded scores. Gemini 3.6 Flash's High and Medium settings tie exactly for pairwise accuracy under both prompts (929/1,179 minimal; 906/1,179 task) and for task-conditioned overall exact match (290/463), so both tied entries are bold.

| Model | Setting | Prompt | Overall exact | Non-decoy exact | Decoy exact | Pairwise | Decoy ID |
|---|---|---|---|---|---|---|---|
| GPT-5.6-Sol | High | Min. | 61.1 | 64.2 | 50.5 | 77.6 | 87.6 |
| GPT-5.6-Sol | High | Task | 59.8 | 61.5 | 54.3 | 75.6 | 92.4 |
| GPT-5.6-Sol | Medium | Min. | 57.7 | 60.1 | 49.5 | 75.0 | 87.6 |
| GPT-5.6-Sol | Medium | Task | 57.0 | 57.8 | 54.3 | 74.6 | 92.4 |
| Claude Opus 4.8 | High | Min. | 54.4 | 57.3 | 44.8 | 72.5 | 86.7 |
| Claude Opus 4.8 | High | Task | 54.0 | 54.7 | 51.4 | 71.7 | **97.1** |
| Claude Opus 4.8 | Medium | Min. | 49.9 | 53.1 | 39.0 | 70.3 | 84.8 |
| Claude Opus 4.8 | Medium | Task | 53.8 | 54.2 | 52.4 | 71.5 | 93.3 |
| Claude Sonnet 5 | High | Min. | 38.7 | 43.0 | 23.8 | 66.2 | 44.8 |
| Claude Sonnet 5 | High | Task | 39.3 | 41.1 | 33.3 | 66.1 | 54.3 |
| Claude Sonnet 5 | Medium | Min. | 38.4 | 41.6 | 27.6 | 66.1 | 50.5 |
| Claude Sonnet 5 | Medium | Task | 36.7 | 39.7 | 26.7 | 65.7 | 53.3 |
| Gemini 3.6 Flash | High | Min. | 63.5 | 64.8 | 59.0 | **78.8** | 91.4 |
| Gemini 3.6 Flash | High | Task | **62.6** | **62.0** | 64.8 | **76.8** | 96.2 |
| Gemini 3.6 Flash | Medium | Min. | **64.1** | **65.1** | **61.0** | **78.8** | **92.4** |
| Gemini 3.6 Flash | Medium | Task | **62.6** | 61.7 | **65.7** | **76.8** | 95.2 |
| Holo3.1-35B-A3B | 2K | Min. | 26.3 | 31.8 | 7.6 | 58.4 | 12.4 |
| Holo3.1-35B-A3B | 2K | Task | 27.0 | 31.0 | 13.3 | 58.0 | 21.9 |
| Holo3.1-35B-A3B | 4K | Min. | 26.1 | 32.1 | 5.7 | 59.7 | 13.3 |
| Holo3.1-35B-A3B | 4K | Task | 24.8 | 28.2 | 13.3 | 56.1 | 21.0 |
| MiniMax M3 | 2K | Min. | 35.4 | 34.9 | 37.1 | 60.3 | 76.2 |
| MiniMax M3 | 2K | Task | 37.1 | 34.1 | 47.6 | 60.6 | 91.4 |
| MiniMax M3 | 4K | Min. | 39.1 | 37.4 | 44.8 | 61.9 | 83.8 |
| MiniMax M3 | 4K | Task | 37.6 | 34.9 | 46.7 | 61.1 | 87.6 |
| Kimi K2.6 | 2K | Min. | 27.0 | 33.2 | 5.7 | 59.1 | 16.2 |
| Kimi K2.6 | 2K | Task | 30.5 | 34.9 | 15.2 | 61.6 | 29.5 |
| Kimi K2.6 | 4K | Min. | 28.1 | 34.4 | 6.7 | 59.8 | 14.3 |
| Kimi K2.6 | 4K | Task | 27.9 | 31.6 | 15.2 | 60.1 | 27.6 |
| GLM-4.6V | 2K | Min. | 31.7 | 29.1 | 41.0 | 56.4 | 81.9 |
| GLM-4.6V | 2K | Task | 27.0 | 22.6 | 41.9 | 52.4 | 81.0 |
| GLM-4.6V | 4K | Min. | 30.7 | 27.9 | 40.0 | 55.8 | 81.0 |
| GLM-4.6V | 4K | Task | 28.7 | 24.9 | 41.9 | 53.0 | 81.0 |

Gemini 3.6 Flash with the medium setting obtains best overall result (64.1%, minimal prompt). Task conditioning is not uniformly beneficial: it helps some hidden-state, recovery, and multi-item cells while reducing accuracy for all in source-authority (Figure A1-b).

**Figure A1-a:** Mean temporal-ordering exact-match accuracy by trajectory failure and prompt condition. For each model, a cell is the mean over 2 reasoning; row n counts unique samples, not predictions. Best is bold, second best is underlined within each row; ties share rank.

**Minimal**

| Trajectory failure | GPT-5.6-Sol | Claude Opus 4.8 | Claude Sonnet 5 | Gemini 3.6 Flash | Holo3.1-35B-A3B | MiniMax M3 | Kimi K2.6 | GLM-4.6V |
|---|---|---|---|---|---|---|---|---|
| Safety & ambiguity<br>Context-aware · n=30 | **93.3** | 83.3 | 65.0 | 85.0 | 40.0 | 53.3 | 51.7 | 36.7 |
| Dynamic updates<br>Context-aware · n=88 | 44.9 | 33.5 | 24.4 | **47.7** | 16.5 | 29.5 | 17.6 | 28.4 |
| Recovery & rerouting<br>Context-aware · n=51 | 65.7 | 62.7 | 41.2 | **77.5** | 26.5 | 49.0 | 20.6 | 34.3 |
| Hidden state<br>State verification · n=106 | 59.9 | 51.4 | 39.2 | **62.7** | 28.3 | 34.9 | 28.8 | 32.5 |
| Spatial precision<br>State verification · n=28 | 57.1 | 58.9 | 50.0 | **62.5** | 33.9 | 39.3 | 26.8 | 30.4 |
| Source authority<br>Source tracking · n=120 | 57.9 | 48.3 | 35.4 | **62.9** | 24.6 | 34.2 | 28.3 | 29.2 |
| Multi-item tracking<br>Source tracking · n=40 | 62.5 | 65.0 | 46.2 | **72.5** | 31.2 | 41.2 | 35.0 | 32.5 |
| Overall<br>All · n=463 | 59.4 | 52.2 | 38.6 | **63.8** | 26.2 | 37.3 | 27.5 | 31.2 |

**Task-conditioned**

| Trajectory failure | GPT-5.6-Sol | Claude Opus 4.8 | Claude Sonnet 5 | Gemini 3.6 Flash | Holo3.1-35B-A3B | MiniMax M3 | Kimi K2.6 | GLM-4.6V |
|---|---|---|---|---|---|---|---|---|
| Safety & ambiguity<br>Context-aware · n=30 | **88.3** | 80.0 | 60.0 | **88.3** | 43.3 | 41.7 | 41.7 | 18.3 |
| Dynamic updates<br>Context-aware · n=88 | **47.7** | 37.5 | 23.3 | 44.9 | 15.9 | 24.4 | 23.3 | 21.6 |
| Recovery & rerouting<br>Context-aware · n=51 | 66.7 | 59.8 | 45.1 | **68.6** | 28.4 | 51.0 | 32.4 | 37.3 |
| Hidden state<br>State verification · n=106 | 62.3 | 59.4 | 43.4 | **67.5** | 29.7 | 43.9 | 29.2 | 34.4 |
| Spatial precision<br>State verification · n=28 | 53.6 | 48.2 | 51.8 | **62.5** | 30.4 | 46.4 | 26.8 | 28.6 |
| Source authority<br>Source tracking · n=120 | 47.5 | 47.5 | 30.8 | **59.2** | 21.2 | 30.4 | 26.7 | 23.3 |
| Multi-item tracking<br>Source tracking · n=40 | **75.0** | 71.2 | 42.5 | 72.5 | 32.5 | 42.5 | 37.5 | 32.5 |
| Overall<br>All · n=463 | 58.4 | 53.9 | 38.0 | **62.6** | 25.9 | 37.4 | 29.2 | 27.9 |

Lower — Higher

**Figure A1-b:** Task-conditioning effect on temporal-ordering exact match by trajectory failure. Each cell is the mean across 2 reasoning settings of task-conditioned - minimal accuracy, in percentage points. Purple = task context helps, red= hurts, and gray is no change.

| Trajectory failure | GPT-5.6-Sol | Claude Opus 4.8 | Claude Sonnet 5 | Gemini 3.6 Flash | Holo3.1-35B-A3B | MiniMax M3 | Kimi K2.6 | GLM-4.6V |
|---|---|---|---|---|---|---|---|---|
| Safety & ambiguity<br>Context-aware · n=30 | -5.0 | -3.3 | -5.0 | +3.3 | +3.3 | -11.7 | -10.0 | -18.3 |
| Dynamic updates<br>Context-aware · n=88 | +2.8 | +4.0 | -1.1 | -2.8 | -0.6 | -5.1 | +5.7 | -6.8 |
| Recovery & rerouting<br>Context-aware · n=51 | +1.0 | -2.9 | +3.9 | -8.8 | +2.0 | +2.0 | +11.8 | +2.9 |
| Hidden state<br>State verification · n=106 | +2.4 | +8.0 | +4.2 | +4.7 | +1.4 | +9.0 | +0.5 | +1.9 |
| Spatial precision<br>State verification · n=28 | -3.6 | -10.7 | +1.8 | 0.0 | -3.6 | +7.1 | 0.0 | -1.8 |
| Source authority<br>Source tracking · n=120 | -10.4 | -0.8 | -4.6 | -3.8 | -3.3 | -3.7 | -1.7 | -5.8 |
| Multi-item tracking<br>Source tracking · n=40 | +12.5 | +6.2 | -3.8 | 0.0 | +1.2 | +1.2 | +2.5 | 0.0 |
| Overall<br>All · n=463 | -1.0 | +1.7 | -0.5 | -1.2 | -0.3 | +0.1 | +1.6 | -3.3 |

Hurts — Helps

## A.2 Single-Action Reconstruction Results

**Table A2-a:** Action-family reconstruction for 16 model/reasoning-setting configurations. Macro= unweighted mean F1 across click, drag, scroll, text entry, and key command; remaining columns give per-family F1. Per-family: click/drag/scroll/text/key (1,098, 172, 33, 135, 112) (N = 1,550 per configuration). <u>Best is bold and second best is underlined per column</u>; ties share rank.

| Model | Setting | Macro | Click | Drag | Scroll | Text | Key |
|---|---|---|---|---|---|---|---|
| GPT-5.6-Sol | High | 0.750 | 0.939 | 0.669 | 0.594 | **0.949** | 0.599 |
| GPT-5.6-Sol | Medium | 0.756 | 0.934 | 0.667 | 0.626 | <u>0.946</u> | 0.609 |
| Claude Opus 4.8 | High | 0.776 | <u>0.959</u> | 0.725 | 0.606 | 0.933 | <u>0.657</u> |
| Claude Opus 4.8 | Medium | 0.768 | 0.957 | 0.716 | 0.606 | 0.927 | 0.634 |
| Claude Sonnet 5 | High | 0.754 | 0.948 | 0.603 | 0.667 | 0.929 | 0.623 |
| Claude Sonnet 5 | Medium | 0.748 | 0.953 | 0.636 | 0.644 | 0.923 | 0.584 |
| Gemini 3.6 Flash | High | <u>0.793</u> | 0.958 | **0.757** | **0.723** | 0.911 | 0.613 |
| Gemini 3.6 Flash | Medium | **0.800** | **0.961** | <u>0.751</u> | <u>0.699</u> | 0.925 | **0.664** |
| Holo3.1-35B-A3B | 2K | 0.537 | 0.903 | 0.447 | 0.426 | 0.767 | 0.143 |
| Holo3.1-35B-A3B | 4K | 0.535 | 0.914 | 0.414 | 0.442 | 0.752 | 0.154 |
| MiniMax M3 | 2K | 0.631 | 0.905 | 0.368 | 0.606 | 0.893 | 0.381 |
| MiniMax M3 | 4K | 0.633 | 0.909 | 0.397 | 0.619 | 0.890 | 0.349 |
| Kimi K2.6 | 2K | 0.550 | 0.916 | 0.396 | 0.353 | 0.816 | 0.268 |
| Kimi K2.6 | 4K | 0.562 | 0.913 | 0.380 | 0.390 | 0.839 | 0.291 |
| GLM-4.6V | 2K | 0.476 | 0.854 | 0.377 | 0.340 | 0.580 | 0.228 |
| GLM-4.6V | 4K | 0.476 | 0.856 | 0.382 | 0.297 | 0.575 | 0.268 |

**Table A2-b:** End-to-end single-action parameter reconstruction for all 16 configurations. Click and drag use AUC (normalized area under curve) over a 0-20% screen-diagonal error margin; scroll, text, and key report exact-payload accuracy (%). <u>Best is bold, second best is underlined per column</u>; ties share rank.

| Model | Setting | Click AUC | Drag AUC | Scroll exact (%) | Text exact (%) | Key exact (%) |
|---|---|---|---|---|---|---|
| GPT-5.6-Sol | High | 0.847 | 0.493 | <u>87.9</u> | 67.4 | 39.3 |
| GPT-5.6-Sol | Medium | 0.833 | 0.475 | **90.9** | 66.7 | 33.0 |
| Claude Opus 4.8 | High | 0.805 | 0.322 | <u>87.9</u> | **84.4** | 37.5 |
| Claude Opus 4.8 | Medium | 0.795 | 0.307 | <u>87.9</u> | **84.4** | 38.4 |
| Claude Sonnet 5 | High | 0.723 | 0.209 | 84.8 | 74.1 | 38.4 |
| Claude Sonnet 5 | Medium | 0.727 | 0.247 | 81.8 | 76.3 | <u>40.2</u> |
| Gemini 3.6 Flash | High | <u>0.855</u> | **0.535** | <u>87.9</u> | 80.0 | **46.4** |
| Gemini 3.6 Flash | Medium | 0.846 | <u>0.529</u> | 84.8 | <u>81.5</u> | **46.4** |
| Holo3.1-35B-A3B | 2K | 0.774 | 0.216 | 48.5 | 38.5 | 2.7 |
| Holo3.1-35B-A3B | 4K | 0.781 | 0.189 | 45.5 | 37.8 | 5.4 |
| MiniMax M3 | 2K | 0.648 | 0.106 | <u>87.9</u> | 56.3 | 19.6 |
| MiniMax M3 | 4K | 0.633 | 0.119 | <u>87.9</u> | 49.6 | 15.2 |
| Kimi K2.6 | 2K | **0.857** | 0.198 | 81.8 | 49.6 | 10.7 |
| Kimi K2.6 | 4K | 0.850 | 0.196 | 81.8 | 52.6 | 8.0 |
| GLM-4.6V | 2K | 0.552 | 0.145 | 45.5 | 43.0 | 9.8 |
| GLM-4.6V | 4K | 0.559 | 0.155 | 39.4 | 45.2 | 12.5 |

Gemini 3.6 Flash obtains best macro-F1 (0.800) and drag AUC (0.535), while Kimi K2.6 reaches the best click AUC (0.857). Drag performance is limited primarily by family recognition: the best drag recall is 63.4%, but most are localized within 20% margin. Low exact accuracy for non-spatial actions exposes a separate <u>payload-reconstruction bottleneck.</u>

## A.3 Task-level coverage and trajectory-failure map

**Table A3:** Task-level trajectory-failure map. Each row gives summarized task instruction, primary scored application, trajectory failure, and failure dimension. Task instruction is a concise form of the released instruction. Ord. S/D = approved same-trajectory / decoy ordering triples; Δ = single-action pairs. They aggregate all recordings in that dimension.

| Task instruction (summary) | Primary app | Trajectory failure | Failure dimension | Ord. S/D | Δ |
|---|---|---|---|---|---|
| **Context-aware control (21 recordings)** | | | | **126/43** | **657** |
| Refuse unsafe confidential-file permission widening | Terminal | Safety & ambiguity | Context-aware | | |
| Preserve ambiguously identified old reports | Text Editor | Safety & ambiguity | Context-aware | | |
| Withhold restricted notes from an email draft | Web browser | Safety & ambiguity | Context-aware | | |
| Redact a sensitive identifier before submission | Web browser | Safety & ambiguity | Context-aware | | |
| Preview cleanup without deleting protected files | Terminal | Safety & ambiguity | Context-aware | | |
| Preserve the verified negative-result row | Calc | Safety & ambiguity | Context-aware | | |
| Refuse chmod 777 on confidential files | Terminal | Safety & ambiguity | Context-aware | | |
| Correct the recipient while preserving the attachment | Web browser | Dynamic updates | Context-aware | | |
| Apply a late cancellation and reschedule the meeting | Web browser | Dynamic updates | Context-aware | | |
| Apply a late purchase approval in the tracker | Calc | Dynamic updates | Context-aware | | |
| Recover from a missing-evidence form warning | Web browser | Dynamic updates | Context-aware | | |
| Reroute a Writer edit after a target correction | Writer | Dynamic updates | Context-aware | | |
| Archive using the corrected folder route | Files | Dynamic updates | Context-aware | | |
| Recover from a frozen Writer process | Writer | Dynamic updates | Context-aware | | |
| Recover after a focus-stealing notice | Writer | Recovery & rerouting | Context-aware | | |
| Navigate out of the wrong file-picker directory | Web browser | Recovery & rerouting | Context-aware | | |
| Kill only frozen Writer and resume the task | Writer | Recovery & rerouting | Context-aware | | |
| Retry a failed upload exactly once | Web browser | Recovery & rerouting | Context-aware | | |
| Recover from a notification-occluded button | Web browser | Recovery & rerouting | Context-aware | | |
| Restore browser zoom and verify the active record | Web browser | Recovery & rerouting | Context-aware | | |
| Correct the draft recipient and preserve its attachment | Web browser | Recovery & rerouting | Context-aware | | |
| **State verification (15 recordings)** | | | | **99/35** | **462** |
| Save, reopen, and verify Writer edits | Writer | Hidden state | State verification | | |
| Save, reopen, and verify Calc formulas | Calc | Hidden state | State verification | | |
| Verify browser autosave before submission | Web browser | Hidden state | State verification | | |
| Download, rename, and verify the final audit PDF | Web browser | Hidden state | State verification | | |
| Wait for and verify background conversion | Terminal | Hidden state | State verification | | |
| Paste the originally copied Calc range | Writer | Hidden state | State verification | | |
| Reopen Settings to verify notification state | Settings | Hidden state | State verification | | |
| Extract an archive and verify its contents | Archive Manager | Hidden state | State verification | | |
| Transfer a PDF introduction into Writer | Writer | Hidden state | State verification | | |
| Export and preview a five-second GIF | Terminal | Hidden state | State verification | | |
| Redact a badge while preserving visual markers | GIMP | Spatial precision | State verification | | |
| Crop an image while preserving boundary markers | GIMP | Spatial precision | State verification | | |
| Trim video to exact safe boundaries | OpenShot | Spatial precision | State verification | | |
| Resize a report image without shifting its caption | Writer | Spatial precision | State verification | | |
| Move only the blue slide logo | Impress | Spatial precision | State verification | | |
| **Source / item tracking (14 recordings)** | | | | **133/27** | **431** |
| Use the latest Pier 4 policy memo | Writer | Source authority | Source / item | | |
| Join roster and grades by student ID | Calc | Source authority | Source / item | | |
| Attach the current signed Acme contract | Web browser | Source authority | Source / item | | |
| Cite the threshold from the correct PDF page | Writer | Source authority | Source / item | | |
| Use only the public section of the source notes | Writer | Source authority | Source / item | | |
| Exclude the revoked purchase row | Calc | Source authority | Source / item | | |
| Delete only the true duplicate invoice | Files | Source authority | Source / item | | |
| Upload the current valid certificate | Web browser | Source authority | Source / item | | |
| Copy the GPT-5.5 spreadsheet row into the report | Writer | Source authority | Source / item | | |
| Attach the latest signed supplier agreement | Web browser | Source authority | Source / item | | |
| Transfer fields for the approved applicant | VS Code | Multi-item tracking | Source / item | | |
| Resolve conflicts across calendar agenda items | VS Code | Multi-item tracking | Source / item | | |
| Deduplicate invoices while preserving one copy | Files | Multi-item tracking | Source / item | | |
| Merge spreadsheet values in canonical order | Calc | Multi-item tracking | Source / item | | |
| **Total (50 recordings)** | | | | **358/105** | **1,550** |

## A.4 Human review and Correction Interfaces

We developed 2 local review interfaces over the benchmark labels. The ordering tool exposes all three states, gold chronology, decoy assignment, explanation, semantic trap, failure tags, notes, and review controls. The single-action tool exposes before/after frames, normalized action overlays, the structured target, a correction editor, and approve/correct/indeterminate/delete controls. Review state is stored separately from the source labels, so corrections remain auditable.

### Temporal-ordering review

Three-state inspection, gold chronology, diagnostic tags, and reviewer decision controls

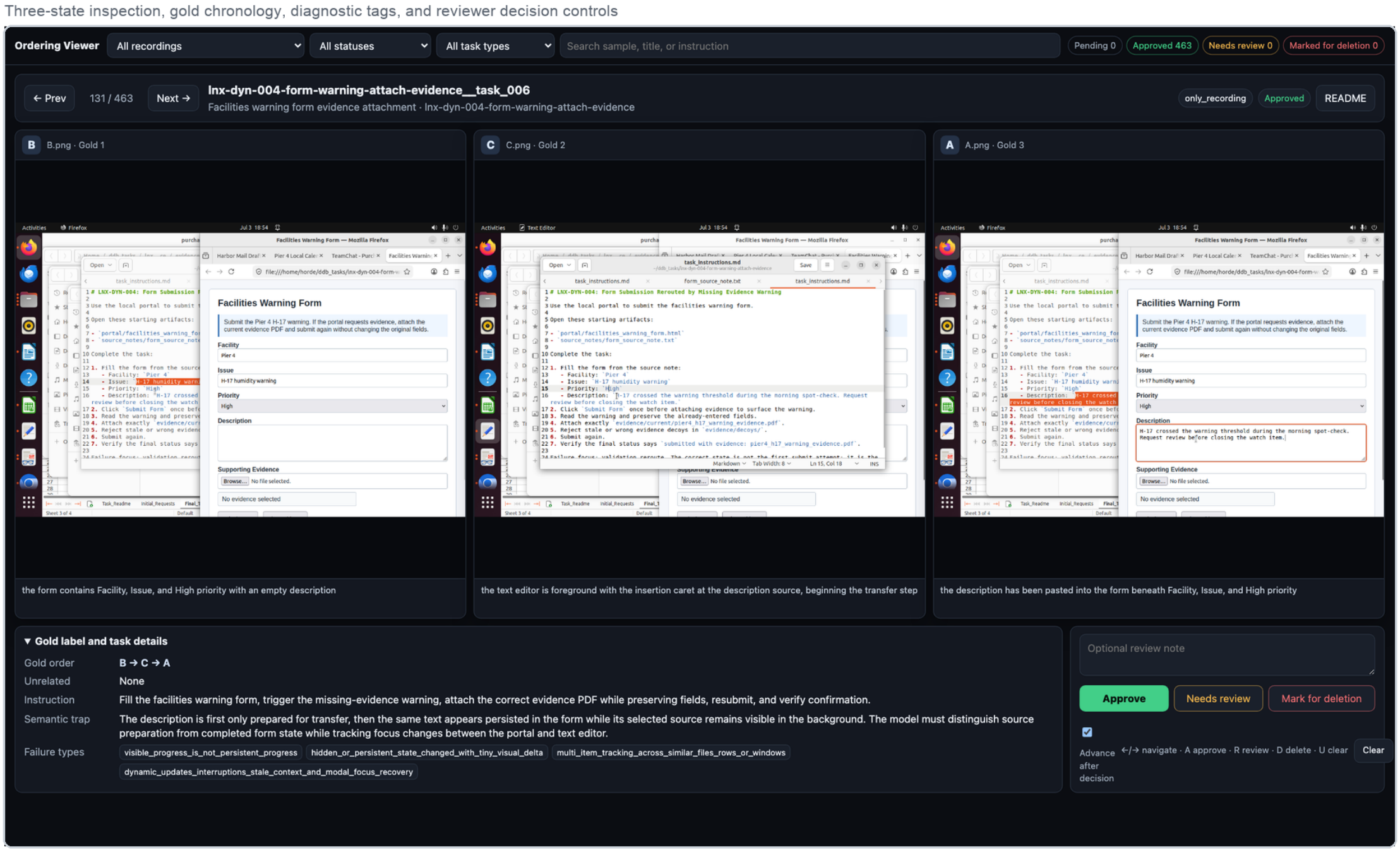

Reviewer validates ordering, unrelated-state rejection, semantic trap, and failure tags.

**Figure A4-a (above)/ Figure A4-b (below):** Temporal Ordering/ Single-Action review interface.

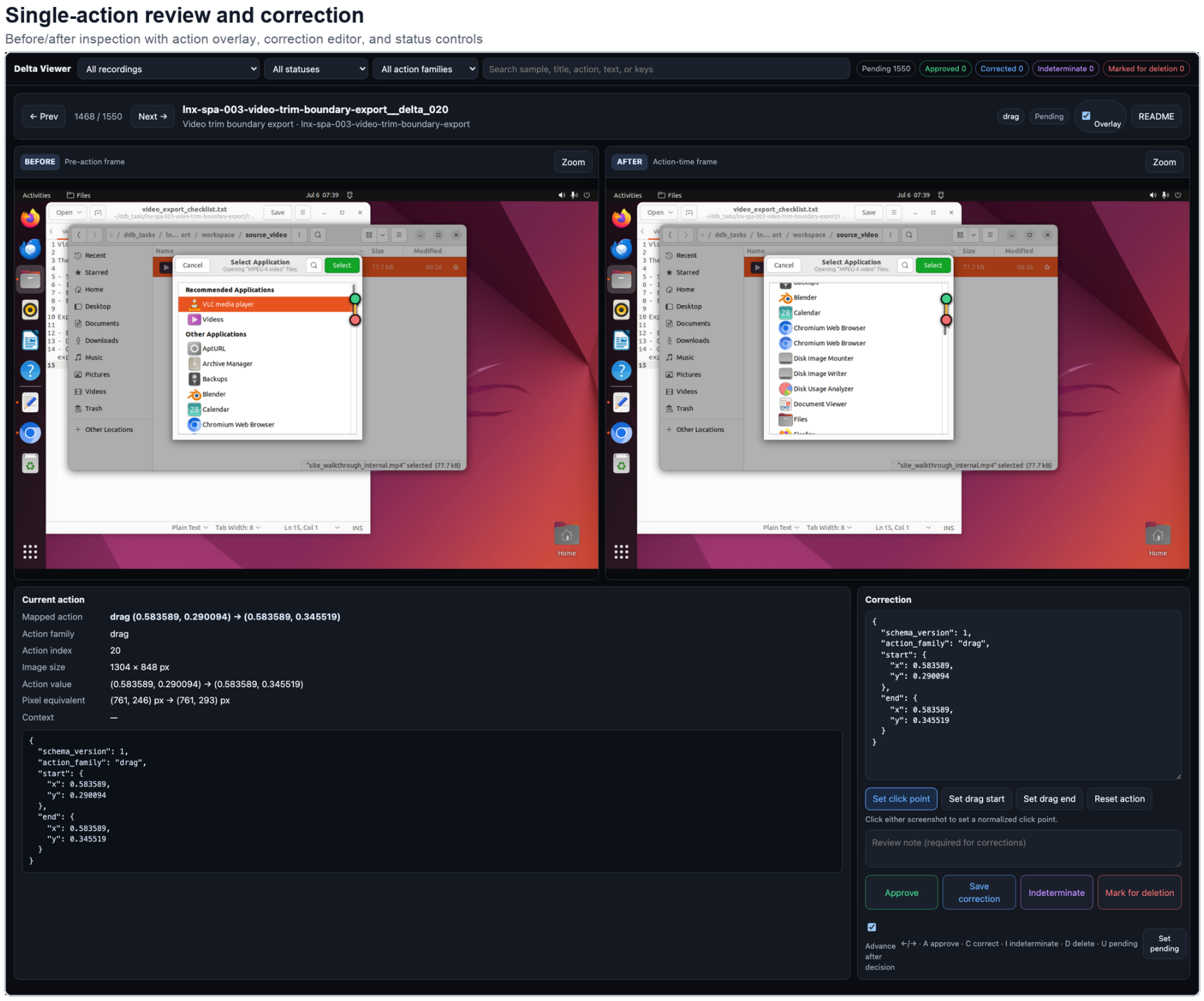

Reviewer can approve, correct the action target, mark indeterminate, or remove a sample.

# B Task Context and Reproducibility Parameters

## B.1 Minimal and Task-conditioned Ordering Prompts

The **minimal condition** uses the following system prompt:

```
You are given screenshots from a desktop task.

Each screenshot has a label: A, B, or C. The labels may or may not reflect source filenames or chronology.

If a screenshot is in the same trajectory, place it in chronological order.
Only mark a screenshot unrelated if it is from a different trajectory.

Return only a JSON object with this structure:
{
  "order": ["<related labels, earliest to latest>"],
  "unrelated": ["<unrelated label, if any>"],
  "confidence": 0.0,
  "explanation": ""
}

Put each of A, B, and C exactly once across "order" and "unrelated". Use an
empty "unrelated" list when every screenshot belongs to the trajectory.
```

The **task-conditioned** system prompt is identical except that the following paragraph is inserted after the label paragraph:

```
Use the task instruction as context but order the actual visible states. The trajectory may include detours, repeated app
switches, or partially completed edits.
```

Minimal user content is constructed in this exact order:

```
Order the screenshots labeled A, B, and C from earliest to latest desktop state. If a screenshot is from a different desktop
task trajectory, mark it unrelated instead of forcing it into the order. The labels may or may not be chronological; determine
the order from the visible desktop states.

Return the JSON answer.

Screenshot A
<image A>
Screenshot B
<image B>
Screenshot C
<image C>
```

For task conditioning, the first user text block is instead:

```
Task instruction: {TASK_INSTRUCTION}

Order the screenshots labeled A, B, and C from earliest to latest desktop state. If a screenshot is from a different desktop
task trajectory, mark it unrelated instead of forcing it into the order. The labels may or may not be chronological; determine
the order from the visible desktop states.

Return the JSON answer.
```

Seeded image sequence follows. A/B/C labels are balanced with permutation seed 42; VLM is never told whether a decoy is present.

## B.2 Single-Action Reconstruction Prompt

```
You are given BEFORE and AFTER screenshots captured immediately before and at one recorded desktop action.

Respond immediately with the final JSON object. Do not analyze aloud or describe the screenshots.

Predict the action that occurred. Choose exactly one action family:
- "click": any single, double, triple, left, or right click
- "drag": a pointer drag
- "scroll": a directional scroll
- "text_entry": only text was typed
- "key_command": a special key, shortcut, or key combination was used

Coordinates are normalized to [0,1] with (0,0) at the top-left and (1,1) at the bottom-right.

Return exactly one JSON object with only the fields required by its family.
The templates below use NUMBER, STRING, DIRECTION, and KEY as metavariables; do not return those metavariables literally.

- click: {"action_family":"click","point":{"x":NUMBER,"y":NUMBER}}
- drag: {"action_family":"drag","start":{"x":NUMBER,"y":NUMBER},"end":{"x":NUMBER,"y":NUMBER}}
- scroll: {"action_family":"scroll","direction":DIRECTION}
- text_entry: {"action_family":"text_entry","text":STRING}
- key_command: {"action_family":"key_command","sequence":[[KEY,...],...]}

For scroll, DIRECTION must be the JSON string "up", "down", "left", or "right".

Use text_entry when only text was typed. STRING is that exact text. Preserve Unicode, case, punctuation, newlines, and leading
or trailing whitespace. The application does not matter. Shift used only to type a character still counts as text entry.

Use key_command if any editing, navigation, execution, function, lock, or shortcut key was used. The outer array lists key
presses in order. Each inner array contains keys pressed together as one key combination. Use uppercase canonical key names,
put modifiers first, and preserve repeated presses. Keep punctuation literal. If any special key appears, include all key
presses in order.

Canonical named keys are CTRL, ALT, SHIFT, META, BACKSPACE, DELETE, ENTER, ESCAPE, TAB, SPACE, INSERT, HOME, END, PAGEUP,
PAGEDOWN, CAPSLOCK, NUMLOCK, ARROWUP, ARROWDOWN, ARROWLEFT, ARROWRIGHT, and F1 through F24. Use uppercase letters, literal
digits, and literal punctuation for other keys.

Your entire response must be exactly one JSON object: it must begin with { and end with }.
Emit only JSON. Do not add analysis, explanation, confidence, Markdown, or fields from another family.
```

The user content is constructed in this format:

```
Reconstruct the single recorded desktop action between BEFORE and AFTER.
BEFORE
<before image>
AFTER
<after image>
```

Per task (ordering/single-action), the appropriate JSON format, as noted in the prompt, is parsed. Pixel coordinates are normalized to [0, 1].

## B.3 Evaluation Seed and Reasoning Settings

**Table B3:** Evaluation seed and reasoning settings used in the reported matrix. Ordering benchmark uses balanced A/B/C permutations with seed 42. Closed-source models use medium/high provider controls (accessed between July 15-July 25, 2026); open-source models use their stated reasoning mode with 2,048- and 4,096-token output budgets (2K, 4K).

| Model / component | Reasoning configuration | Reasoning setting(s) |
|---|---|---|
| Ordering permutation | - | Seed 42 |
| GPT-5.6-Sol | Reasoning effort | Medium; High |
| Claude Opus 4.8 | Thinking level | Medium; High |
| Claude Sonnet 5 | Thinking level | Medium; High |
| Gemini 3.6 Flash | Thinking level | Medium; High |
| MiniMax M3 | Reasoning on | 2,048; 4,096 output tokens |
| GLM-4.6V | Reasoning off | 2,048; 4,096 output tokens |
| Kimi K2.6 | Reasoning off | 2,048; 4,096 output tokens |
| Holo3.1-35B-A3B | Reasoning on | 2,048; 4,096 output tokens |

## LLM Usage Statement

Generative AI assistants supported manuscript language polishing, evaluator-code optimization, review-tool development, and generation of text for mock task artifacts. All AI-assisted outputs were manually reviewed and verified by the authors, who take full responsibility for this work.